\documentclass[10pt,twocolumn]{article}

\usepackage[a4paper,margin=0.7in]{geometry}
\usepackage[T1]{fontenc}
\usepackage{lmodern}

\usepackage{graphicx}
\usepackage{booktabs}
\usepackage{amsmath}
\usepackage{amssymb}
\usepackage{microtype}
\usepackage{cite}
\usepackage[hidelinks]{hyperref}
\usepackage{xcolor}
\usepackage{titling}

\newcommand{\Description}[1]{}

\pretitle{\centering\Large\bfseries}
\posttitle{\par\vskip 0.5em}
\title{
\textbf{Understanding Temporal Semantic Stability in Open-Vocabulary\\
UAV Perception through Metric 3D Fusion}
}

\author{
Saurbh Singh Jamwal\\
Department of Computer Science and Engineering\\
Indian Institute of Technology Bombay\\
\texttt{saurbh@cse.iitb.ac.in}
}

\date{}

\begin{document}
\raggedbottom

\maketitle

\begin{abstract}

Recent open-vocabulary segmentation models have significantly advanced semantic perception for UAVs, for aerial and remote-sensing imagery, yet predictions from moving platforms can remain temporally inconsistent across repeated observations of the same physical scene. While semantic segmentation is typically evaluated on single images, temporal reliability remains less explored despite its importance for long-horizon robotic perception. In this work, we investigate temporal semantic stability by associating frame-wise predictions with persistent world-space locations through metric 3D fusion. We introduce a voxel-level evaluation framework that jointly characterises final semantic agreement, Semantic Belief Drift (SBD), Observation Persistence (OP), and semantic uncertainty from accumulated evidence. Experiments on UAVid-3D-Scenes reveal substantial frame-wise semantic variation and show that high aggregate world-space agreement can overstate temporal stability when many locations have limited repeated-observation support. Persistence-stratified analysis shows that recurrent voxels expose substantially greater semantic disagreement, while higher-persistence voxels exhibit lower direct consecutive semantic-evidence divergence. This behaviour is observed across two segmentation backbones and remains consistent under variations in voxel resolution, geometric association, and temporal sampling density. Overall, conditions that reduce world-space recurrence can increase apparent aggregate stability, demonstrating that semantic consistency must be interpreted together with observation support. Our findings establish observation persistence as an essential conditioning variable for evaluating long-horizon semantic reliability and provide a framework for analysing temporal semantic behaviour beyond image-level prediction. Code and supplementary material are publicly available at \href{https://github.com/SaiyanSam/TeSS-UAV}{\textcolor{blue}{project repository}}.

\end{abstract}


\vspace{0.5em}

\section{Introduction}

Open-vocabulary semantic segmentation has become increasingly important for UAV perception, enabling aerial platforms to assign semantic meaning using flexible task-specific vocabularies rather than a single fixed closed-set taxonomy. Such capabilities are useful for environmental monitoring, infrastructure inspection, semantic navigation, and long-horizon robotic perception. However, predictions from moving UAV platforms can exhibit temporal instability across consecutive frames or repeated viewpoints, reducing the reliability of semantic information used by downstream robotic systems.

Most semantic segmentation methods are evaluated using image-level metrics such as pixel accuracy and mean Intersection-over-Union, where each frame is assessed independently. This provides limited insight into long-horizon UAV perception, where a moving platform repeatedly observes the same physical locations from changing viewpoints, scales, and image positions. Existing video segmentation methods primarily address temporal coherence in image space, while metric-semantic mapping systems typically assume that semantic observations are sufficiently reliable to accumulate into persistent representations.

This motivates the central question of this work: \emph{when semantic predictions from multiple UAV observations are associated with the same physical location, how does the corresponding semantic belief evolve over time?} Raw frame-wise variation alone cannot answer this question, since same-pixel comparisons on a moving UAV also reflect camera-induced displacement. Moreover, a physical location may eventually exhibit high semantic agreement despite substantial changes in its accumulated belief during repeated observation.

We study this problem using metric 3D fusion as an analysis mechanism rather than as a proposed semantic mapping system. Frame-wise semantic predictions are projected using depth and camera pose into a shared world-space representation, allowing observations from changing viewpoints to be associated with persistent voxel locations. We then characterise temporal semantic behaviour using complementary measures of Voxel Semantic Stability (VSS), Semantic Belief Drift (SBD), Observation Persistence (OP), and semantic entropy, which describe agreement, belief evolution, repeated-observation support, and residual uncertainty rather than semantic correctness.

Experiments on UAVid-3D-Scenes show that semantic stability must be interpreted jointly with observation persistence. Sparsely observed voxels can exhibit deceptively high apparent agreement, whereas repeated observations expose semantic disagreement and residual uncertainty. Changing voxel resolution and geometric association alters recurrence, while denser temporal sampling exposes disagreement that sparse observations can hide. Across these analyses, conditions that reduce world-space recurrence can increase apparent aggregate stability.

We focus primarily on predominantly static scene regions, where repeated observations can be associated with persistent spatial locations. Dynamic objects require motion-aware or object-level association because fixed voxels cannot preserve object identity over time.

The main contributions of this work are:
\begin{itemize}

\item We present a systematic study of temporal semantic instability in UAV perception using an open-vocabulary segmentation backbone instantiated with a task-specific vocabulary, distinguishing raw frame-wise variation from semantic evolution at persistent physical locations.

\item We use metric 3D fusion as an evaluation mechanism for world-space temporal analysis across multiple segmentation backbones and changing UAV viewpoints.

\item We introduce a voxel-level framework comprising VSS, SBD, OP, and semantic entropy to characterise final agreement, belief evolution, repeated observation, and residual uncertainty.

\item We show through persistence-stratified, voxel-resolution, geometric-perturbation, and temporal-sampling analyses that sparse or degraded associations can inflate aggregate agreement, while recurrent voxels expose semantic disagreement hidden by aggregate stability alone.

\end{itemize}

\section{Related Work}

\paragraph{Semantic Segmentation for UAV and Remote-Sensing Perception}

Semantic segmentation is a fundamental component of UAV perception, enabling aerial platforms to interpret scene structure for applications such as environmental monitoring, infrastructure inspection, semantic navigation, and autonomous flight. Datasets such as UAVid~\cite{lyu2020uavid} and DeepGlobe~\cite{demir2018deepglobe} have accelerated progress in semantic understanding for aerial and remote-sensing imagery. Recent segmentation models include transformer-based architectures such as SegFormer~\cite{xie2021segformer}, which improve dense prediction through efficient multi-scale feature modeling, and foundation-model-based approaches such as the Segment Anything Model (SAM)~\cite{kirillov2023segment}, which demonstrate strong zero-shot mask generation capabilities. For remote-sensing imagery, Wang \textit{et al.}~\cite{wang2023samrs} introduced SAMRS, while Li \textit{et al.}~\cite{li2025segearthov} proposed SegEarth-OV, a training-free open-vocabulary segmentation framework designed for Earth observation imagery.

Although these methods have improved semantic prediction quality and category generalisation, they are commonly evaluated using image-level metrics that assess each frame independently. Such evaluations provide limited insight into temporal semantic reliability when models are deployed on moving UAV platforms. In practice, viewpoint changes, scale variation, occlusions, and open-vocabulary ambiguity can cause semantic labels to fluctuate across repeated observations of the same physical scene. This paper focuses on studying this temporal semantic instability, rather than proposing a new segmentation model.

\paragraph{Temporal Consistency in Sequential Perception}

Temporal consistency has been widely studied in video understanding and video segmentation to reduce frame-to-frame prediction variability. Bertasius and Torresani~\cite{bertasius2020maskprop} proposed mask propagation techniques for jointly classifying, segmenting, and tracking object instances across video sequences. Cheng \textit{et al.}~\cite{cheng2021rethinking} introduced STCN, a memory-based framework that improves temporal coherence through space-time feature propagation, while Li \textit{et al.}~\cite{li2022videoknet} proposed Video K-Net for unified video segmentation with temporal feature association.

These approaches primarily improve temporal coherence in image space, where semantic predictions remain associated with pixel locations in individual frames. However, UAV perception systems operate under camera motion and repeatedly observe the same physical locations from changing viewpoints. Image-space temporal consistency therefore does not directly measure whether semantic evidence assigned to the same world-space location is stable over time. In contrast, our work studies temporal semantic instability after geometric alignment, allowing semantic behavior to be analysed with respect to persistent physical locations rather than frame-specific pixels.

\paragraph{Metric-Semantic Mapping and World-Space Representations}

Metric semantic mapping systems combine geometric reconstruction with semantic prediction to build structured scene representations for robotic reasoning. SemanticFusion~\cite{mccormac2017semanticfusion} integrated per-frame semantic segmentation with dense SLAM to accumulate semantic predictions in a globally consistent 3D map. PanopticFusion~\cite{narita2019panopticfusion} extended this direction by reconstructing semantic and instance-level scene representations, while Kimera~\cite{rosinol2020kimera,rosinol2021kimera} introduced metric-semantic and object-aware scene graph representations for higher-level spatial reasoning.

Semantic information has also been used in UAV autonomy and planning. Bartolomei \textit{et al.}~\cite{bartolomei2020perceptionaware} incorporated semantic segmentation into perception-aware path planning, and later studied semantic-aware active perception using deep reinforcement learning~\cite{bartolomei2021semanticaware}. Yue \textit{et al.}~\cite{yue2024semanticnav} proposed semantic-driven visual navigation for UAVs, while Xu \textit{et al.}~\cite{xu2022visionaided} combined visual perception with gradient-based trajectory optimisation for UAV navigation and obstacle avoidance.

These works demonstrate the importance of semantic information for robotic perception and decision making. However, their primary goal is to construct or exploit semantic representations, whereas our goal is to analyse the temporal reliability of the semantic observations that such systems may consume. Many metric-semantic systems implicitly assume that frame-wise semantic predictions are reliable enough to be accumulated after geometric alignment. In this paper, we instead ask how stable those semantic predictions remain when repeatedly associated with the same physical locations.

Existing research has made substantial progress in semantic segmentation, video-level temporal consistency, and metric-semantic mapping. However, the temporal instability of open-vocabulary semantic predictions in UAV perception remains insufficiently characterised from a world-space perspective. Image-level segmentation metrics measure semantic correctness per frame, video segmentation methods focus on image-space coherence, and semantic mapping systems emphasise representation construction. In contrast, this work uses metric 3D fusion as an evaluation mechanism to study temporal semantic instability in UAV perception. By accumulating semantic evidence in persistent voxels, we quantify semantic agreement, semantic belief drift, observation persistence, and semantic uncertainty after geometric alignment. This provides a complementary evaluation perspective for long-horizon UAV perception beyond conventional image-level accuracy metrics.

\section{Semantic Stability Analysis Framework}

\subsection{Temporal Semantic Stability}

We define \emph{temporal semantic stability} as the consistency of semantic predictions assigned to the same physical location under repeated observations. Unlike frame-wise consistency, which compares predictions in image space, we evaluate stability after semantic observations are geometrically associated in a shared world-space representation. This is important for UAV perception, where camera motion and viewpoint changes cause the same scene element to appear at different image locations over time.

Metric 3D fusion is therefore used as an evaluation substrate rather than as a proposed semantic mapping system. Persistent world-space associations allow us to analyse semantic agreement, belief evolution, observation persistence, and uncertainty under repeated observation.

\subsection{Semantic Evidence and World-Space Association}

Given an image $I_t$, a segmentation model produces a dense label map
\[
S_t : \Omega \rightarrow \mathcal{C},
\]
where $\Omega$ is the image domain and $\mathcal{C}$ is the semantic class set. Discrete predictions are represented as one-hot evidence vectors $\mathbf{s}_t(u,v)\in[0,1]^K$, where $K=|\mathcal{C}|$; soft class scores can instead be used directly.

Given depth $d_t(u,v)$ and camera intrinsics $(f_x,f_y,c_x,c_y)$, each pixel is back-projected as
\[
\mathbf{x}_c =
d_t(u,v)
\begin{bmatrix}
(u-c_x)/f_x \\
(v-c_y)/f_y \\
1
\end{bmatrix}.
\]
Using the camera-to-world transformation $\mathbf{T}_{wc}$,
\[
\begin{bmatrix}
\mathbf{x}_w\\
1
\end{bmatrix}
=
\mathbf{T}_{wc}
\begin{bmatrix}
\mathbf{x}_c\\
1
\end{bmatrix},
\]
yielding the world-space semantic observation
\[
\mathbf{o}_w=(\mathbf{x}_w,\mathbf{s}_t(u,v)).
\]

\subsection{Persistent Voxel Evidence}

World-space observations are discretised at resolution $r$ as
\[
\mathbf{q}
=
\left\lfloor
\frac{\mathbf{x}_w}{r}
\right\rfloor.
\]
To prevent multiple pixels from one frame being interpreted as repeated temporal evidence, observations assigned to voxel $\mathbf{q}$ within frame $t$ are averaged:
\[
\bar{\mathbf{s}}_{t,\mathbf{q}}
=
\frac{1}{|\mathcal{A}_t(\mathbf{q})|}
\sum_{\mathbf{s}_i\in\mathcal{A}_t(\mathbf{q})}
\mathbf{s}_i,
\]
where $\mathcal{A}_t(\mathbf{q})$ is the set of observations from frame $t$ assigned to $\mathbf{q}$. Each voxel therefore contributes at most one evidence vector per frame.

For a voxel observed in $T_{\mathbf{q}}$ frames, its temporal history and accumulated evidence are
\[
\mathcal{H}(\mathbf{q})
=
\left\{
\bar{\mathbf{s}}_{1,\mathbf{q}},
\ldots,
\bar{\mathbf{s}}_{T_{\mathbf{q}},\mathbf{q}}
\right\},
\qquad
\mathbf{e}(\mathbf{q})
=
\sum_{i=1}^{T_{\mathbf{q}}}
\bar{\mathbf{s}}_{i,\mathbf{q}},
\]
with normalized belief
\[
\mathbf{p}_{\mathbf{q}}
=
\frac{\mathbf{e}(\mathbf{q})}
{\sum_k e_k(\mathbf{q})}.
\]

\subsection{Voxel-Level Stability Metrics}

We characterise each voxel using four complementary measures.

\paragraph{Voxel Semantic Stability.}
VSS measures the concentration of accumulated evidence on the dominant class:
\[
\mathrm{VSS}(\mathbf{q})
=
\frac{\max_k e_k(\mathbf{q})}
{\sum_k e_k(\mathbf{q})}.
\]
Higher VSS indicates stronger final semantic agreement. However, $\mathrm{OP}=1$ voxels have maximal VSS by construction and provide no evidence of temporal consistency.

\paragraph{Semantic Belief Drift.}
The accumulated belief after the first $t$ observations is
\[
\mathbf{p}_{\mathbf{q}}^{(t)}
=
\frac{
\sum_{i=1}^{t}\bar{\mathbf{s}}_{i,\mathbf{q}}
}{
\sum_k
\left[
\sum_{i=1}^{t}\bar{\mathbf{s}}_{i,\mathbf{q}}
\right]_k
}.
\]
For $T_{\mathbf{q}}\geq2$, Semantic Belief Drift is
\[
\mathrm{SBD}(\mathbf{q})
=
\frac{1}{T_{\mathbf{q}}-1}
\sum_{t=2}^{T_{\mathbf{q}}}
D_{\mathrm{JS}}
\left(
\mathbf{p}_{\mathbf{q}}^{(t)},
\mathbf{p}_{\mathbf{q}}^{(t-1)}
\right),
\]
measuring the average change in accumulated belief between successive observations. Because these beliefs are cumulative, each additional observation has progressively smaller influence as the history grows; we therefore interpret SBD as a descriptor of belief evolution rather than treating decreasing SBD alone as evidence of temporal stabilization.

\paragraph{Observation Persistence.}
Observation Persistence measures repeated observation support:
\[
\mathrm{OP}(\mathbf{q})=T_{\mathbf{q}}.
\]
In particular, $\mathrm{OP}=1$ indicates that temporal consistency cannot be assessed, while higher OP provides progressively stronger temporal evidence.

\paragraph{Semantic Entropy.}
Residual uncertainty is measured using normalized entropy:
\[
H(\mathbf{q})
=
\frac{
-\sum_k p_{\mathbf{q},k}\log p_{\mathbf{q},k}
}{
\log K
}.
\]
Low entropy indicates concentrated evidence, while high entropy indicates competing semantic interpretations.

VSS, SBD, OP, and entropy respectively capture final agreement, belief evolution, observation support, and residual uncertainty. They are temporal diagnostics rather than measures of semantic correctness: repeated observations can consistently support an incorrect class. We therefore interpret them jointly, with OP conditioning whether apparent stability is supported by repeated observation.

\section{Experiments}

\subsection{Experimental Setup}

Experiments are conducted on the public UAVid-3D-Scenes extension of UAVid~\cite{lyu2020uavid,TanDepth2025}, which provides undistorted RGB imagery, metric depth maps, camera intrinsics, and camera poses. We evaluate five representative scenes: \textit{scene0}, \textit{scene1}, \textit{scene3}, \textit{scene5}, and \textit{scene8}. For the main multi-scene evaluation, 20 frames are used per scene with the same spatial sampling stride of four pixels. Differences in the number of usable projected samples therefore arise primarily from scene-dependent valid-depth coverage rather than different frame or spatial sampling.

SegEarth-OV~\cite{li2025segearthov} serves as the primary open-vocabulary segmentation backbone and is instantiated with an explicit eight-class UAVid vocabulary. SegFormer~\cite{xie2021segformer} is additionally evaluated under a matched world-space protocol. For persistence-stratified cross-backbone comparison only, predictions are mapped to a shared {building, road, vegetation} taxonomy. For frame-wise analysis, we additionally evaluate SAM~\cite{kirillov2023segment} using class-agnostic masks with a simple area-based pseudo-semantic heuristic, together with raw SAM proposals.

Raw frame-wise flicker is measured as the percentage of valid pixels whose semantic labels change at the same image coordinates between consecutive sampled frames. For world-space analysis, semantic predictions are associated using depth, camera intrinsics, and poses, and accumulated into persistent voxels. We evaluate these voxels using Voxel Semantic Stability (VSS), Semantic Belief Drift (SBD), Observation Persistence (OP), and normalised semantic entropy. Unless otherwise stated, the voxel resolution is $0.2\,\mathrm{m}$.

In addition to the main five-scene evaluation, we analyse stability as a function of observation persistence and evaluate sensitivity to voxel resolution, geometric perturbations, and temporal sampling density. These experiments examine how repeated-observation support and world-space association affect the interpretation of apparent semantic stability.

\begin{figure*}[t]
    \centering
    \includegraphics[width=\textwidth]{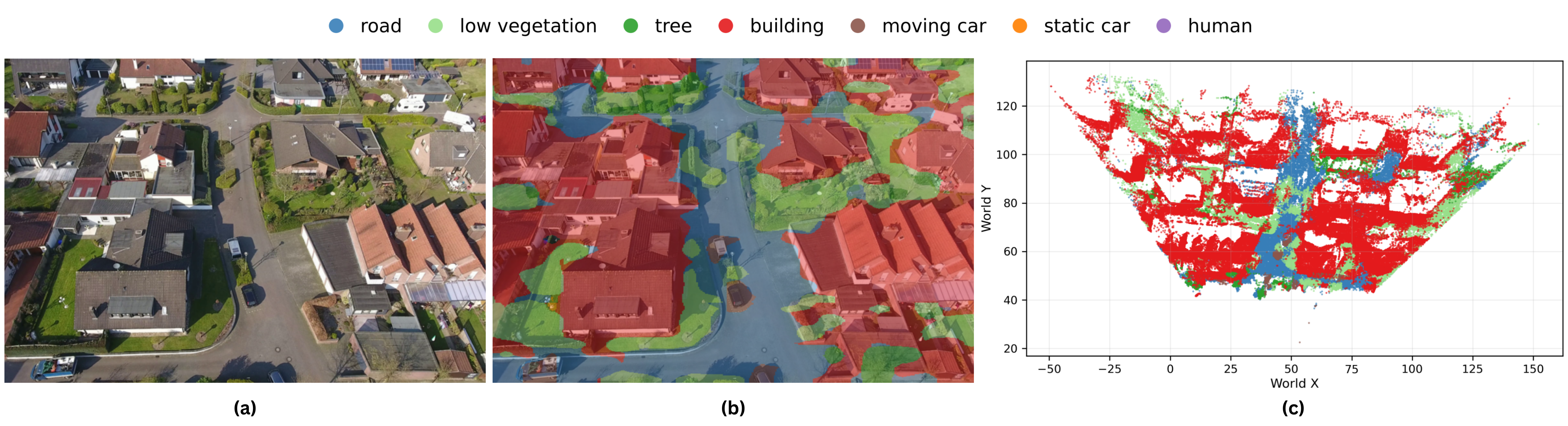}
    \caption{World-space semantic fusion for a representative UAVid-3D-Scenes sequence: (a) RGB frame, (b) SegEarth-OV prediction, and (c) fused semantic evidence used for temporal stability analysis.}
    \Description{
    Three-panel visualization showing an aerial RGB image, its color-coded semantic segmentation, and the corresponding fused semantic representation in a top-down world-space view. The fused view contains spatially distributed colored regions representing semantic observations accumulated across the UAV sequence.
    }
    \label{fig:global_fusion}
\end{figure*}

\subsection{Frame-wise Temporal Semantic Flicker}

We first quantify raw frame-wise semantic variation directly in image space. Flicker is measured as the percentage of valid pixels whose semantic labels change at the same image coordinates between consecutive sampled frames. We evaluate SAM with heuristic semantic assignment, SegFormer, and SegEarth-OV, with raw SAM proposal variation reported as a reference. Because the UAV camera is moving, this same-pixel comparison can also reflect camera-induced image displacement. Table~\ref{tab:flicker_baselines} summarises this result.

\begin{table}[t]
\centering
\caption{Raw same-pixel frame-wise semantic flicker on the 20-frame scene0 evaluation sequence. Lower values indicate less image-space variation.}
\label{tab:flicker_baselines}
\begin{tabular}{lc}
\toprule
Method & Mean Flicker (\%) \\
\midrule
SAM + heuristic & 29.47 \\
Raw SAM proposals & 21.64 \\
SegFormer & 25.39 \\
SegEarth-OV & 15.73 \\
\bottomrule
\end{tabular}
\end{table}

All evaluated methods exhibit substantial raw image-space variation. SAM with heuristic semantic assignment produces 29.47\% flicker, SegFormer 25.39\%, and SegEarth-OV 15.73\%. To quantify the contribution of camera motion, we additionally warp SegEarth-OV predictions between consecutive frames using the released metric depth, camera intrinsics, and poses. Across the five scenes, mean scene-level flicker decreases from 30.25\% under the same-pixel comparison to 26.24\% after geometric alignment over 1.189 million valid correspondences. Thus, ego-motion explains part, but not all, of the observed frame-wise variation.

We therefore treat raw image-space flicker as a motivational diagnostic rather than a direct measure of world-space semantic instability, and next analyse temporally repeated observations at persistent voxel locations.

\subsection{World-Space Evidence Fusion}

Frame-wise predictions are projected using depth, camera intrinsics, and poses and accumulated into persistent voxels as described in Sec.~3. Figure~\ref{fig:global_fusion} illustrates the resulting world-space representation. All 20 sampled frames are successfully processed in each of the five scenes without skipped frames. The fused representation associates observations from changing viewpoints with common world-space locations and serves as the evaluation substrate for the subsequent temporal analysis.

\subsection{Multi-Scene Voxel-Level Stability}
\label{multi_scene}

We next evaluate temporal semantic stability in world space across the five UAVid-3D-Scenes sequences. For each scene, frame-wise SegEarth-OV predictions are projected into metric 3D space and accumulated into persistent voxels. Each voxel stores frame-level semantic evidence observations, from which we compute Voxel Semantic Stability (VSS), Semantic Belief Drift (SBD), Observation Persistence (OP), and normalised semantic entropy.

Table~\ref{tab:multiscene_results} reports resulting multi-scene stability statistics. Across all scenes, the framework constructs 705,902 persistent voxels from 1,353,903 sampled world-space semantic points. The macro-average VSS is 0.957, indicating high aggregate final semantic agreement after metric association. However, the macro-average scene-level mean OP is only 1.51, indicating that many voxels receive limited repeated-observation support. Aggregate VSS must therefore be interpreted jointly with observation persistence rather than as direct evidence that all reconstructed locations are temporally stable.

\begin{table*}[t]
\centering
\caption{Multi-scene world-space semantic stability analysis using persistent voxel evidence. VSS measures final semantic agreement, SBD measures semantic belief drift, OP denotes observation persistence, and entropy measures final semantic uncertainty.}
\label{tab:multiscene_results}
\small
\setlength{\tabcolsep}{8pt}
\begin{tabular}{lrrrrrr}
\toprule
Scene & Voxels & Sampled Points & Mean OP & VSS $\uparrow$ & SBD $\downarrow$ & Entropy $\downarrow$ \\
\midrule
scene0 & 306,957 & 637,068 & 2.08 & 0.951 & 0.0466 & 0.0395 \\
scene1 & 320,725 & 624,234 & 1.95 & 0.908 & 0.0949 & 0.0758 \\
scene3 & 27,904  & 32,999  & 1.18 & 0.981 & 0.0714 & 0.0133 \\
scene5 & 23,862  & 27,433  & 1.15 & 0.978 & 0.1007 & 0.0155 \\
scene8 & 26,454  & 32,169  & 1.22 & 0.966 & 0.1091 & 0.0239 \\
\midrule
Total / Macro Mean & 705,902 & 1,353,903 & 1.51 & 0.957 & 0.0845 & 0.0336 \\
\bottomrule
\end{tabular}
\end{table*}

The results reveal two complementary trends. First, final semantic agreement is high at the aggregate level, with VSS values above 0.90 in every scene. However, high VSS alone does not establish temporal consistency: voxels observed only once have maximal VSS by construction, and the relatively low mean OP in several scenes indicates that such sparsely observed locations contribute substantially to the aggregate statistics. This motivates explicitly separating apparent stability due to limited observation from stability supported by repeated observations.

Second, SBD remains non-zero across all scenes, showing that accumulated semantic beliefs can evolve even when their final evidence is concentrated on a dominant class. For example, \textit{scene5} and \textit{scene8} achieve VSS values of 0.978 and 0.966, respectively, while exhibiting higher SBD than \textit{scene0}. Thus, final semantic agreement and temporal belief evolution capture distinct aspects of semantic behaviour.

The scene-level differences further highlight the role of observation coverage. All scenes use the same frame count and spatial sampling, but usable depth coverage is 97.51\% and 95.52\% in \textit{scene0} and \textit{scene1}, compared with only 5.11\%, 4.28\%, and 4.93\% in \textit{scene3}, \textit{scene5}, and \textit{scene8}. The resulting differences in projected samples and mean OP therefore primarily reflect scene-dependent valid-depth coverage. We next stratify the analysis by observation persistence to determine how apparent stability changes when evaluation is restricted to recurrent voxels.

\subsection{Analysis of Semantic Stability Metrics}

The multi-scene results show high aggregate final semantic agreement, but also reveal that stability cannot be interpreted independently of repeated-observation support. We therefore analyse the proposed metrics in greater detail, examining how semantic behaviour varies with observation persistence, belief evolution, uncertainty, and semantic class. In particular, we distinguish sparsely observed locations from recurrent voxels for which temporal stability can be meaningfully assessed.

\subsubsection{Semantic Agreement and Belief Drift}

\begin{figure*}[t]
    \centering
    \includegraphics[width=0.88\textwidth]{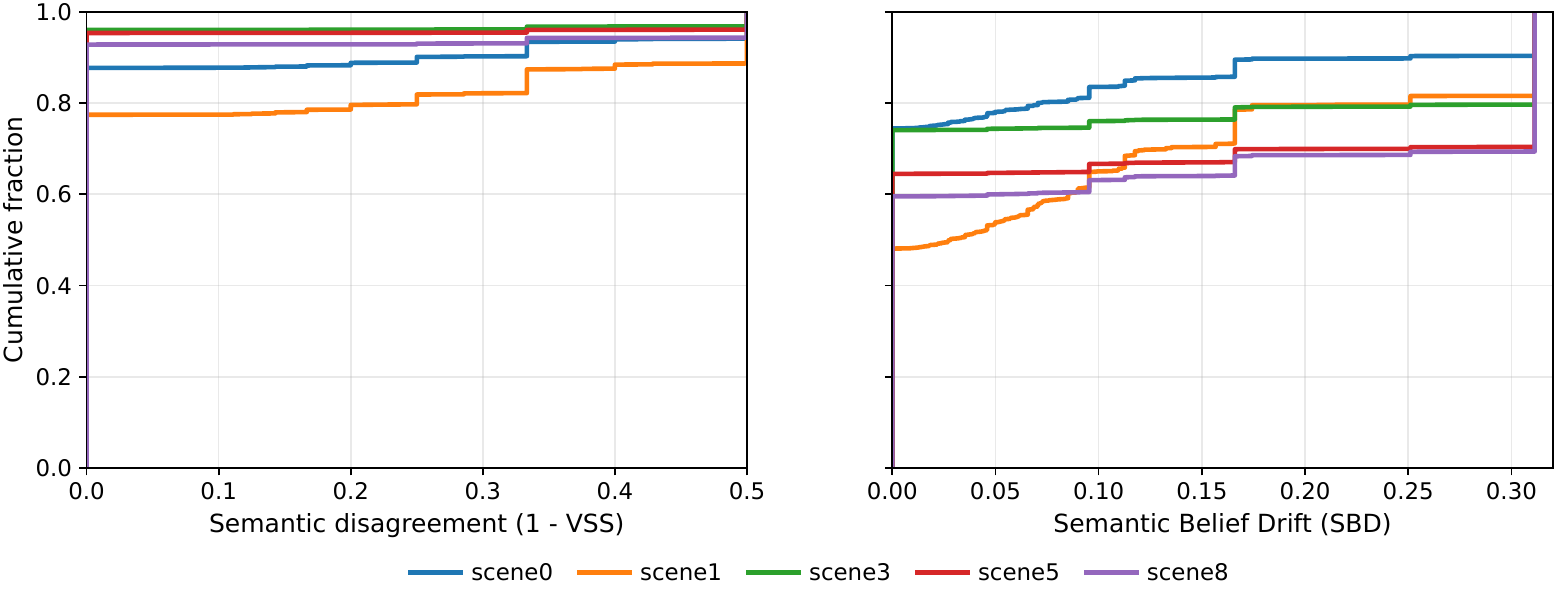}
    \caption{Voxel-level semantic instability across UAVid-3D-Scenes sequences: empirical cumulative distributions of (left) semantic disagreement ($1-\mathrm{VSS}$) and (right) Semantic Belief Drift (SBD), where lower values indicate greater stability.}
    \Description{Two empirical cumulative distribution plots showing voxel-level semantic instability across the evaluated scenes. The left plot shows semantic disagreement, defined as one minus VSS, and the right plot shows Semantic Belief Drift. Multiple curves represent the evaluated UAVid-3D-Scenes sequences and are concentrated toward low instability values.}
    \label{fig:disagreement_sbd_ecdf}
\end{figure*}

Voxel Semantic Stability (VSS) and Semantic Belief Drift (SBD) capture complementary aspects of temporal semantic behaviour. VSS measures the final concentration of accumulated semantic evidence on the dominant class, whereas SBD measures how much the accumulated belief changes as observations arrive. A voxel can therefore exhibit high final agreement while still undergoing non-zero belief evolution before reaching its final accumulated semantic distribution.

Figure~\ref{fig:disagreement_sbd_ecdf} shows that aggregate disagreement and drift are concentrated near zero. However, aggregate VSS must be interpreted carefully because a voxel observed only once necessarily has VSS$=1$ and zero entropy, while SBD is undefined. High aggregate agreement can therefore arise either from genuinely consistent repeated observations or simply from insufficient temporal evidence. This motivates explicitly conditioning the subsequent analysis on observation persistence.

\subsubsection{Observation Persistence}

Observation Persistence (OP) measures the number of frame-level semantic observations supporting a voxel and determines how much temporal evidence is available for stability analysis. We therefore stratify the SegEarth-OV voxel population under the shared three-class evaluation protocol, which is also used in the cross-backbone comparison.

\begin{table}[t]
\centering
\caption{Persistence-stratified semantic stability for SegEarth-OV under the shared three-class taxonomy at $0.2\,\mathrm{m}$ voxel resolution. Fractions are computed over the corresponding reconstructed voxel population.}
\label{tab:persistence_stratified}
\small
\setlength{\tabcolsep}{4pt}
\begin{tabular}{lrrrr}
\toprule
Persistence & Voxels (\%) & VSS $\uparrow$ & SBD $\downarrow$ & Entropy $\downarrow$ \\
\midrule
OP$=1$ & 57.89 & 1.000 & -- & 0.000 \\
OP$=2$ & 20.32 & 0.873 & 0.0790 & 0.1602 \\
$3\leq$OP$\leq5$ & 18.02 & 0.869 & 0.0454 & 0.2285 \\
OP$>5$ & 3.77 & 0.869 & 0.0216 & 0.2530 \\
\bottomrule
\end{tabular}
\end{table}

Table~\ref{tab:persistence_stratified} highlights the gap between apparent aggregate stability and stability supported by repeated observations. About 57.9\% of voxels are observed once and thus have VSS$=1$ and zero entropy by construction, whereas recurrent voxels have VSS near $0.87$, revealing disagreement hidden at OP$=1$.

A complementary trend appears in SBD. Mean cumulative SBD decreases from $0.0790$ at OP$=2$ to $0.0454$ for $3\leq\mathrm{OP}\leq5$ and $0.0216$ for OP$>5$. However, in a separate control on the original eight-class evaluation, permuting each voxel's temporal evidence produces nearly identical cumulative SBD for OP$\geq3$ ($0.0497$ original vs. $0.0500$ shuffled), showing that part of this decrease arises from cumulative averaging. In the same control, direct Jensen--Shannon divergence between consecutive evidence vectors decreases from $0.3027$ at OP$=2$ to $0.2748$ for $3\leq\mathrm{OP}\leq5$ and $0.2368$ for OP$>5$, indicating that higher-persistence voxels are associated with lower frame-to-frame semantic variation.

This distinction is central to interpreting world-space temporal stability. OP indicates whether sufficient repeated evidence exists for VSS, SBD, and entropy to support a meaningful temporal interpretation; we therefore report persistence alongside stability metrics throughout the remaining experiments.

\begin{figure*}[t]
    \centering
    \includegraphics[width=\textwidth]{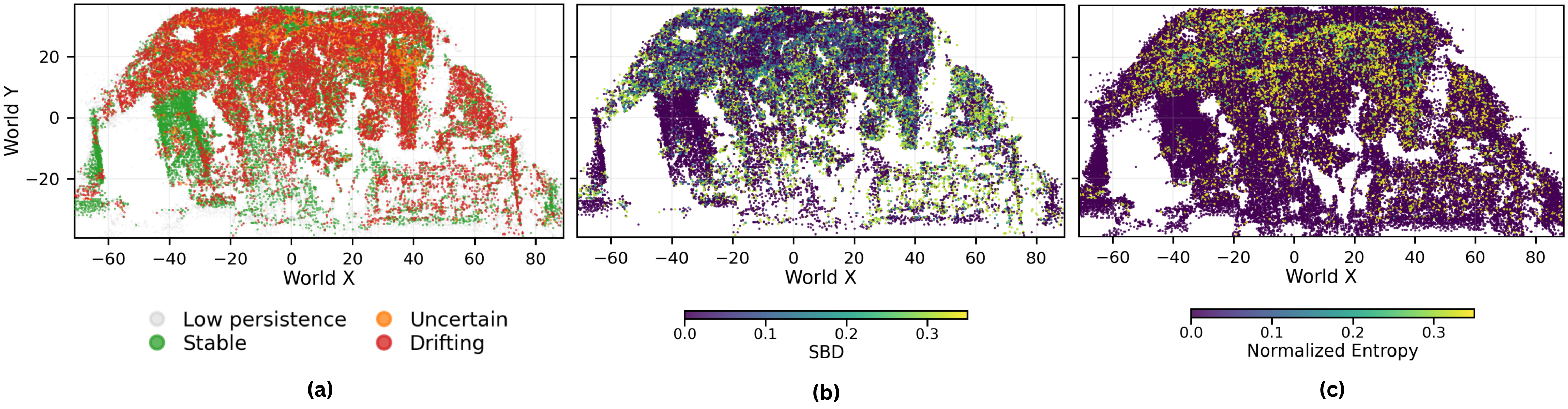}
    \caption{Voxel-level temporal semantic stability for \textit{scene1}: (a) stability categories, (b) Semantic Belief Drift, and (c) semantic uncertainty, showing spatially localised drift and ambiguity.}
    \Description{
    Three top-down world-space visualizations of the same reconstructed UAV scene. The panels show categorical semantic stability, Semantic Belief Drift, and semantic uncertainty, respectively. Most reconstructed regions exhibit low drift and uncertainty, with higher-instability regions appearing spatially localized.
    }
    \label{fig:qualitative_stability}
\end{figure*}

\subsubsection{Semantic Uncertainty}

Semantic entropy measures ambiguity in the final accumulated voxel belief. Its interpretation is also strongly coupled to persistence. Singly observed voxels have zero entropy by construction, whereas entropy increases to $0.1602$ at OP$=2$, $0.2285$ for $3\leq\mathrm{OP}\leq5$, and $0.2530$ for OP$>5$.

This increase should not be interpreted as evidence that repeated observation makes semantic perception less stable. Instead, additional observations expose disagreement that cannot exist in a single-observation belief. Cumulative SBD also decreases with persistence, although the permutation control shows that this trend is partly induced by cumulative averaging. Entropy and SBD therefore describe different properties: entropy measures semantic diversity retained in the accumulated evidence, while SBD describes the evolution of the accumulated belief.

Together with VSS, these results show that final semantic agreement, accumulated uncertainty, and belief evolution capture distinct aspects of recurrent semantic behaviour. A voxel may retain substantial disagreement among observations even when its final accumulated belief is relatively concentrated.

\subsubsection{Class-wise Stability}

We further analyse temporal semantic behaviour by grouping voxels according to their dominant semantic class, allowing us to examine whether instability is concentrated in particular semantic categories. Table~\ref{tab:classwise_stability} reports class-wise voxel counts, VSS, SBD, and entropy across all evaluated scenes.

\begin{table}[t]
\centering
\caption{Class-wise semantic stability across all scenes; rare classes should be interpreted cautiously.}
\label{tab:classwise_stability}
\small
\setlength{\tabcolsep}{4pt}
\begin{tabular}{lrrrr}
\toprule
Class & Voxels & VSS $\uparrow$ & SBD $\downarrow$ & Entropy $\downarrow$ \\
\midrule
building & 286,565 & 0.939 & 0.0626 & 0.0464 \\
road & 207,895 & 0.915 & 0.0913 & 0.0685 \\
tree & 97,988 & 0.929 & 0.0851 & 0.0612 \\
low vegetation & 102,703 & 0.961 & 0.0393 & 0.0375 \\
moving car & 10,564 & 0.960 & 0.0498 & 0.0403 \\
\bottomrule
\end{tabular}
\end{table}

The dominant classes in the reconstructed voxel space are building, road, tree, and low vegetation. Road and tree exhibit higher SBD and entropy than low vegetation, indicating greater variability in their accumulated semantic evidence. Such differences may arise from viewpoint changes, semantic boundaries, or ambiguity between visually similar surface classes.

Class-wise statistics must nevertheless be interpreted jointly with observation persistence. Moving-car voxels have lower recurrence than major static classes (mean OP $=1.66$; $17.81\%$ have OP$\geq3$), so their aggregate VSS of $0.960$ is partly driven by sparsely observed voxels. Restricting the analysis to OP$\geq3$ reduces moving-car VSS to $0.7729$, comparable to tree ($0.7719$) and below road, low vegetation, and building. We therefore avoid interpreting moving cars as intrinsically more stable than static classes. Static car and human contain too few recurrent voxels for reliable comparison.

\subsection{Cross-Backbone Temporal Stability}
\label{cross_backbone}

We evaluate SegEarth-OV and SegFormer under the same five-scene world-space protocol using a shared three-class taxonomy. Under this mapping, SegEarth retains $97.59\%$ of valid-depth semantic samples, whereas SegFormer retains $68.35\%$, because predictions outside the shared taxonomy are ignored before projection. SegFormer therefore produces fewer recurrent voxels: $15.29\%$ have OP$\geq3$ compared with $21.79\%$ for SegEarth-OV. Aggregate stability is nevertheless higher for SegFormer (VSS $0.9848$, SBD $0.0380$) than for SegEarth-OV (VSS $0.9657$, SBD $0.0661$).

The difference remains after conditioning on recurrence. For OP$\geq3$, SegFormer achieves VSS $0.9410$ and SBD $0.0186$, versus $0.8692$ and $0.0413$ for SegEarth-OV. Thus, recurrence differences reflect backbone-dependent semantic coverage, while recurrent voxels still exhibit different temporal behaviour across backbones.

These results illustrate why aggregate temporal metrics alone can be misleading in cross-backbone comparisons. Here, OP should be interpreted as recurrence of retained shared-taxonomy evidence rather than as a purely geometric property.

\subsection{Sensitivity to World-Space Association}
\label{sensitivity_analysis}

We next test whether the persistence-aware interpretation remains consistent under changes that affect world-space association. Varying voxel resolution from $0.1$ to $0.4\,\mathrm{m}$ increases the fraction of recurrent voxels (OP$\geq3$) from $3.90\%$ to $42.61\%$, while aggregate VSS decreases from $0.9824$ to $0.9274$. Fine discretisation therefore produces apparently higher stability partly by fragmenting repeated observations across more voxels, whereas coarser discretisation increases recurrence but also semantic mixing. The default $0.2\,\mathrm{m}$ setting provides an intermediate trade-off.

Controlled geometric perturbations produce the complementary effect. With $0.3\%$ relative depth noise, OP$\geq3$ decreases from $22.12\%$ to $9.81\%$, while aggregate VSS increases from $0.9569$ to $0.9777$. This apparent improvement reflects reduced recurrent association rather than greater semantic consistency: recurrent-voxel SBD changes only from $0.0497$ to $0.0541$. Similar behaviour is observed under mild pose perturbations.

Finally, we uniformly sample 20, 40, and 80 frames over the same trajectory extent in \textit{scene0} and \textit{scene1}. Increasing sampling density raises OP$\geq3$ from $1.61\%$ to $18.99\%$ while aggregate VSS decreases from $0.9805$ to $0.9400$. In contrast, recurrent-voxel VSS changes only from $0.8544$ to $0.8446$, while SBD decreases from $0.0579$ to $0.0512$. Denser sampling therefore exposes disagreement that sparse observations cannot reveal without fundamentally changing the behaviour of recurrent locations.

Across all three studies, conditions that reduce recurrence can increase apparent aggregate stability, whereas stronger repeated-observation support exposes semantic disagreement. These results show that apparent stability depends not only on semantic predictions but also on how consistently observations are associated in world space. This reinforces the central finding that VSS and related stability measures must be interpreted jointly with OP.

\subsection{Qualitative Analysis}

We complement the quantitative evaluation with qualitative visualisations of voxel-level semantic behaviour after frame-wise predictions are geometrically associated in world space. The goal is not to demonstrate semantic mapping performance, but to illustrate the spatial distribution of repeated-observation stability, belief drift, and semantic uncertainty.

Figure~\ref{fig:qualitative_stability} shows representative top-down visualisations. Voxels are categorised according to their temporal semantic behaviour: recurrent stable voxels have high final semantic agreement under repeated observation, drifting voxels exhibit elevated Semantic Belief Drift, uncertain voxels contain ambiguous accumulated semantic evidence, and low-persistence voxels lack sufficient repeated observations for reliable temporal assessment. The resulting patterns show that measurable drift and uncertainty are spatially localised rather than uniformly distributed across the reconstructed scene.

The qualitative results reinforce the persistence-aware interpretation of the quantitative analysis. Regions with sufficient repeated observations can exhibit either consistent semantic evidence or localised disagreement and belief evolution, while low-persistence regions must be distinguished from genuinely stable recurrent locations. The visualisation therefore illustrates why VSS and SBD should be interpreted jointly with OP and entropy when analysing temporal semantic reliability in world space.

\section{Limitations and Future Work}

While the analysis provides insight into temporal semantic instability, the proposed metrics characterise temporal behaviour rather than absolute correctness. A model can repeatedly assign the same incorrect label and still obtain high VSS and low SBD. For the evaluated sequences, only one aligned semantic ground-truth frame is available, preventing meaningful temporal 3D mIoU or correctness correlation. The framework should therefore complement conventional accuracy metrics.

World-space stability also depends on geometric association. Pose and depth errors can reduce recurrence and alter the voxel population over which stability is measured, sometimes increasing aggregate VSS simply because fewer observations are associated. Our analysis uses the released UAVid-3D-Scenes geometry, whose coverage varies substantially across scenes; more realistic localization, calibration, and depth-error models remain to be studied.

The experiments remain limited to a small set of UAVid-3D-Scenes trajectories, two segmentation backbones, and predominantly static outdoor environments. SegEarth-OV is instantiated with an explicit eight-class UAVid vocabulary, so the present study does not evaluate unrestricted open-set discovery. Recurrence may also change under different flight patterns, sensing configurations, or dynamic environments, where fixed voxel association cannot preserve object identity.

Future work will extend the analysis to additional models, datasets, and longer trajectories, and investigate motion-aware association for dynamic scenes. A further direction is to move from diagnosing temporal semantic instability to designing persistent semantic memory mechanisms that use repeated, geometrically aligned evidence to decide when semantic beliefs should be reinforced, revised, or corrected over time.

\section{Conclusion}

We presented a world-space framework for analysing temporal semantic stability in UAV perception by associating frame-wise semantic predictions with persistent voxel locations. The framework jointly characterises semantic agreement, belief evolution, observation persistence, and uncertainty using VSS, SBD, OP, and semantic entropy. Experiments on UAVid-3D-Scenes show that aggregate semantic consistency can be misleading in limited repeated-observation support, making observation persistence essential for interpreting temporal reliability. Overall, the analysis provides a complementary perspective to image-level evaluation for studying long-horizon semantic behaviour in robotic platforms.

\bibliographystyle{unsrt}
\bibliography{references}

\end{document}